\documentstyle[ijcai95]{article}
\input{epsf}

\newcommand{\systemname}[1]{{\sc #1}}
\newcommand{\muc}[1]{\mbox{MUC-#1}}

\newcommand{\object}[1]{{\tt #1}}
\newcommand{\slot}[1]{{\tt #1}}
\newcommand{\slotvalue}[1]{{\tt #1}}
\newcommand{\feature}[1]{{\small #1}}

\title{\bf Using Decision Trees\\
           for Coreference Resolution\thanks{
This research was supported by NSF Grant no. EEC-9209623,
State/\-Industry/\-University Cooperative Research on Intelligent
Information Retrieval, Digital Equipment Corporation and the
National Center for Automated Information Research.
}}

\author{Joseph F. McCarthy \and Wendy G. Lehnert\\
        Department of Computer Science\\
        University of Massachusetts\\
        Amherst, MA  01003-4610\\
        Email:  \{jmccarthy,lehnert\}@cs.umass.edu}

\begin{document}

\maketitle

\thispagestyle{myheadings}
\renewcommand{\thepage}{\empty}
\markright{To appear in Proceedings of the
           Fourteenth International Joint Conference on
           Artificial Intelligence (IJCAI '95)}

\begin{abstract}

This paper describes \systemname{resolve}, a system that uses decision
trees to learn how to classify coreferent phrases in the domain of
business joint ventures.  An experiment is presented in which the
performance of \systemname{resolve}\ is compared to the performance of
a manually engineered set of rules for the same task.  The results
show that decision trees achieve higher performance than the rules in
two of three evaluation metrics developed for the coreference task.
In addition to achieving better performance than the rules,
\systemname{resolve}\ provides a framework that facilitates the
exploration of the types of knowledge that are useful for solving the
coreference problem.


\end{abstract}

\section{Introduction}
\label{introduction}

The goal of an {\em Information Extraction} (IE) system is to identify
information of interest from a collection of texts.  Within a
particular text, objects of interest are often referenced in different
places and in different ways.  One of the many challenges facing an IE
system is to determine which references refer to which objects.  This
problem can be recast as a classification problem:  given two
references, do they refer to the same object or different objects.

The Message Understanding Conferences (MUCs)
\cite{muc3:intro:overview,muc4:intro:overview,muc5:intro:overview} and
the Tipster Project \cite{tipster:intro:overview} helped both to
define the information extraction task and to push the technology of
IE systems.  Each of these evaluation efforts provided a corpus of
news articles about a domain, a specification of the relevant
information that was to be extracted from each article, the output
representation of that information, and a set of key templates
representing the information extracted from each article by human
readers.\footnote{The \muc{5} evaluation actually included 4 different
domains, but most participants were required to select only one.}  For
the final evaluations, participating systems were given a set of blind
texts and their output was scored against the key templates to
determine how much of the relevant information they were able to
extract.

The sentence analyzers used in many of these systems have shown
significant improvement over the past several years.  However, the
discourse processing capabilities of these systems, particularly their
coreference resolution components, have often been cited as weak areas
\cite{muc5:system:unisys,muc4:site:usc,muc4:site:mitre}.

The IE systems developed at UMass
\cite{muc3:system:umass,muc4:system:umass,muc5:system:umass} also
displayed weak coreference resolution capabilities.  Each of these
systems used a set of manually engineered rules to resolve some
obvious types of coreference, but they tended to be very conservative,
i.e., they only considered phrases to be coreferent if there was
overwhelming evidence in support of that hypothesis.  One of the
problems with these coreference resolution components was figuring out
which features of the phrases to look at when determining coreference.
Another, related set of problems was determining how to combine
positive and negative evidence into individual rules and then how to
order the rule set.  A third problem area was the accumulation of
errors at that late stage of processing, e.g., from incorrectly
delimited sentences, incorrect part-of-speech tags, and other sentence
analysis errors.

In an effort to address these problems, a new approach to coreference
resolution was begun after the \muc{5} evaluation: a system named
\systemname{resolve}\ was created to build decision trees that can be
used to classify pairs of phrases as coreferent or not coreferent.
The errors generated by the sentence analyzer were eliminated by using
a special tool -- the Coreference Marking Interface, or
\systemname{cmi}\ -- to extract a set of phrases from the \muc{5}
English Joint Venture (EJV) corpus.\footnote{The \muc{5} EJV corpus is
a collection of news articles, written in English, that describe
business joint ventures, i.e., associations of two or more entities
(companies, governments or people) created for the purpose of owning
and/or developing a project together.}  In order to minimize the
difficulties involved with creating and maintaining complex sets of
rules, a machine learning approach was adopted, in which a decision
tree determines the order and relative weight of different pieces of
evidence.

\systemname{Resolve}\ used the C4.5 decision tree system
\cite{quinlan:93:c4.5} to learn how to classify coreferent phrases for
the experiments reported in this paper.  C4.5 was chosen primarily due
to its ease of use and its widespread acceptance; however,
\systemname{resolve}\ can use any learning system that uses feature
vectors composed of attribute-value pairs.

\section{Decision Trees vs. Rules}
\label{experiment}

An experiment was conducted to compare the performance of the decision
trees generated by \systemname{resolve}\ with the performance of manually
engineered rules used for coreference classification in the
UMass/\-Hughes \muc{5} IE system.  A set of references, along with the
coreference links among these, were extracted from a group of texts
via \systemname{cmi}.  All possible pairings of references from each text were
generated, and these pairings were used to create a set of feature
vectors used by \systemname{resolve}.  The pairings that contained coreferent
phrases formed positive instances, while those that contained two
non-coreferent phrases formed negative instances.  \systemname{resolve}\ was
then
iteratively trained and tested on different partitions of this set of
feature vectors.

The data structure used in discourse processing by the UMass/\-Hughes
\muc{5} IE system was the {\em memory token}, which converted the case
frame output from the \systemname{circus}\ sentence analyzer
\cite{lehnert:91:circus} into a more system-independent
representation.  Prior to coreference processing, each memory token
contained one noun phrase, one or more lexical patterns encompassing
that phrase, part-of-speech tags, semantic features, and information
that was inferred from either the phrase or the context in which the
phrase was found.  This inferred information included the type of
object referenced by the phrase, any name or location substring
contained in the phrase, and some domain-specific information such as
whether the phrase was a joint venture parent (one of the entities who
formed a joint venture) or joint venture child (the joint venture
company itself).  The references marked via \systemname{cmi}\ were converted
into a
memory token representation in order to test the performance of the
\muc{5} system's coreference module.

\subsection{Data}
\label{data}

The articles in the EJV corpus describe business joint ventures among
two or more entities (companies, governments and/or people).  The task
definition provided for \muc{5} required IE systems to extract
information about the entities involved, the relationships among these
entities, the facilities associated with the joint venture, the
products or services offered by the joint venture, its capitalization
and revenue projections, and a variety of other related information.
Since the entities involved in these joint ventures were the main
focus of most of these articles, references to entities were much more
numerous than references to other types of object classes, e.g.,
people.  Therefore, \object{entity} references were selected as the
focus of the experiments reported in this paper.

\systemname{Cmi}\ is a graphical user interface that permits the user
to mark phrases in a text; for each phrase, the user can indicate the
object(s) with which the phrase is coreferent and some additional
information about the phrase that can be inferred either from the
phrase itself or its local context.  This additional information is
parameterized and can be modified easily for use in different domains.
The data used in this experiment was based on a set of phrases
extracted using \systemname{cmi}.

As an example, consider the following sentence, from text 0970 from
the \muc{5} EJV corpus:

\begin{quote}
\begin{footnotesize}
\raggedright
\underline{\bf FAMILYMART CO.} OF
\underline{\bf SEIBU SAISON GROUP} WILL OPEN
\\
\underline{A CONVENIENCE STORE}
\underline{IN TAIPEI}
\underline{FRIDAY}
\\
IN
\underline{\bf A JOINT VENTURE} WITH
\underline{\bf TAIWAN'S LARGEST CAR DEALER},
\\
THE COMPANY SAID WEDNESDAY.
\end{footnotesize}
\end{quote}

The phrases underlined in this sentence contain relevant information
that must be extracted by an IE system.\footnote{Note that the phrase
``THE COMPANY'' in the last clause of the sentence is not considered
relevant, since it contributes no information required for the \muc{5}
task -- the determination of who is announcing a joint venture or when
the announcement was made are not relevant pieces of information.
Therefore, this phrase was not marked for use in the experiment.}  The
phrases in {\bf boldface} refer to \object{entity} objects that are
important to the \muc{5} task.  As an example of the types of
information collected about each phrase, consider the first phrase in
the sentence:

\begin{small}
\begin{tabbing}
(\=:string ``FAMILYMART CO.''\\
 \>:slots (\=ENTITY\\
 \>        \>(name ``FAMILYMART CO.'')\\
 \>        \>(type COMPANY)\\
 \>        \>(relationship JV-PARENT CHILD)))\\
\end{tabbing}
\end{small}

Information collected about each phrase includes the string itself,
the character position of the string in the source text (not shown),
the index of the sentence within which the string is found (also not
shown), and some slot information that can be inferred from either the
string itself or its local context -- the same kind of information
that was contained in the memory tokens used by the \muc{5} system.
In this example, the \slot{name} of the \object{entity} and the fact
that it is a \slotvalue{company} entity can both be inferred from the
string itself.  The fact that Familymart Company plans to open a store
in ``A JOINT VENTURE'' with another entity is considered adequate
evidence that the company is the parent of a joint venture
(\slotvalue{jv-parent}); the fact that the sentence contains the
pattern ``{\em company-name-1} OF {\em company-name-2\/}'' is evidence
that {\em company-name-1}, in this case Familymart Co., is a
subsidiary (\slotvalue{child}) of {\em company-name-2}, in this case
Seibu Saison Group.

A second example of output from \systemname{cmi}\ can be seen below, where
\slot{nationality} information has been extracted from the reference
to the car dealer:

\begin{small}
\begin{tabbing}
(\=:string ``TAIWAN'S LARGEST CAR DEALER''\\
 \>:slots (\=ENTITY\\
 \>        \>(type COMPANY)\\
 \>        \>(relationship JV-PARENT)\\
 \>        \>(nationality ``Taiwan (COUNTRY)'')))\\
\end{tabbing}
\end{small}

In principle, much of the information gathered about a particular
string could be found automatically: there are numerous proper name
recognizer programs, programs that extract location information, and
sentence analyzers that can infer relationship information -- any
system that exhibited good performance in \muc{5} must be good at
inferring such relationships.

For the purposes of our experiment, however, this information was
specified by a user via \systemname{cmi}.  The primary motivation for
this was to minimize the noise in the data; coreference resolution
often occurs at a late processing stage in an IE system, and earlier
errors such as incorrect part-of-speech tags, incorrectly delimited
sentences and semantic tagging errors can create significant noise for
a coreference classifier.

\systemname{Cmi}\ was used to mark references to a variety of relevant
object types (\object{entity}, \object{facility}, \object{person} and
\object{product-or-service}) in 50 randomly selected
texts.\footnote{In order to make things manageable for
\systemname{cmi}\ annotator, the size of the texts was limited to 2KB,
however the majority of texts in the EJV domain fall into this
category.}  Since references to \object{entity} objects were most
numerous, this was the object class chosen for the experiment.  In the
50 texts, 472 references to a total of 205 \object{entity} objects
were marked using \systemname{cmi}.

Some phrases are {\em multireferent}, i.e., they refer to more than
one object.  These multireferent phrases pose difficulties for
classification, since it means that some phrases will be coreferent
with other phrases in the text that have distinct referents.  Thus for
a set of phrase pairs which share a given phrase, more than one pair
would be classified as a positive instance of coreference.  Further
complications are created for evaluating the performance of a
coreference system when multireferent phrases are included in the data
(see Section \ref{evaluation}).  To simplify the initial
experiments reported here, multireferent phrases were excluded from the
data set.  The capability to handle such phrases will be incorporated
in a later version of \systemname{resolve}.

\subsection{Rules used in the \muc{5} System}
\label{rules}

The coreference module of the UMass/\-Hughes \muc{5} IE system was
designed to minimize false positives, i.e., minimize the likelihood
that two phrases that were not coreferent would be labeled coreferent.
This design decision was based on the assumption that false positive
errors, resulting in the merging of non-coreferent phrases in the
final system output, would harm system performance more than false
negative errors, which would result in coreferent phrases showing up
in distinct objects in the system output.  This rather conservative
approach to coreference was shared by a number of MUC system
developers
\cite{muc4:site:sri,muc4:system:bbn}, though not all
\cite{muc3:addendum:discourse}.

Another factor influencing the coreference module was the short time
allotted to developing and testing this system component.  Since
coreference resolution was a late stage in processing, upstream
components had to be stabilized before serious development could take
place on coreference.  Several late-stage components were being
developed in parallel, so it is difficult to assess the time devoted
exclusively to developing the coreference module, but we estimate
it was two person-weeks.

The rules used to determine whether two phrases (represented as memory
tokens) were coreferent in the \muc{5} system are shown in Table
\ref{table:muc5-rules}.  Following the policy of minimizing false
positives, whenever none of the rules fired, the system classified the
pair of tokens as not coreferent.

\begin{table}
\centering
\begin{small}
\begin{tabular}{|ll|}
\hline
IF   & both tokens come from the same {\em trigger family} \\
THEN & they are not coreferent. \\
\hline
IF   & each token comes from a different {\em partition} \\
THEN & they are not coreferent. \\
\hline
IF   & both tokens contain a common phrase \\
THEN & they are coreferent. \\
\hline
IF   & both tokens refer to joint ventures \\
THEN & they are coreferent. \\
\hline
IF   & both tokens contain the same company name \\
THEN & they are coreferent. \\
\hline
IF   & one token contains an alias of the other \\
THEN & they are coreferent. \\
\hline
IF   & only one token refers to a joint venture \\
THEN & they are not coreferent. \\
\hline
IF   & each token contains different company names \\
THEN & they are not coreferent. \\
\hline
\end{tabular}
\end{small}
\caption{The \muc{5} system's coreference rules.}
\label{table:muc5-rules}
\end{table}

The UMass/\-Hughes \muc{5} IE system used a variety of mechanisms to
identify phrases referring to joint ventures (the entity formed by two
or more parent entities for some particular business purpose), to
identify company names within a phrase (if they exist), and to
determine whether one phrase was an alias (an abbreviation or
shortened form), as well as the ability to identify trigger
families\footnote{A trigger family is a set of phrases all {\em
triggered} off the same word, e.g., a subject and direct object joined
by the same verb phrase.} and partitions\footnote{A partition is a
portion of the text that is focusing on the same main topic.  For the
\muc{5} system, distinct partitions were recognized only for texts
that had bulleted items, as one might see in a news summary of the
days headlines.  Most texts thus had a single partition.} in the text.

One of the many difficulties in developing the rule set for
coreference classification was in ordering the rules.  Several
different orderings were tested during the development period, and the
order shown above was the ordering of the rule set used for final
evaluation.  This difficulty in rule ordering was one of the
motivations behind using a machine learning approach -- we wanted to
develop a system that could {\em learn} how to combine the positive
and negative evidence.

\subsection{Features Used By RESOLVE}
\label{features}

A decision tree requires data to be represented by feature
vectors, i.e., vectors of attribute/value pairs.  For the task of
coreference classification, references were paired up, and features
were extracted from the pair of references as well as from the
individual references themselves.  Since this experiment involved a
comparison between \systemname{resolve}\ and a manually engineered rule set,
the
features used in this experiment were based on the antecedents of the
coreference rules used in the UMass/ Hughes \muc{5} IE system.

For example, Table \ref{table:ejv-instance} shows a feature vector
that represents the pairing of the phrases ``FAMILYMART CO.'' and
``TAIWAN'S LARGEST CAR DEALER''.  Since the two phrases are not
coreferent, this represents a negative instance.

\begin{table}
\centering
\begin{small}
\begin{tabular}{|l|c|l|c|}
\hline
\multicolumn{2}{|c|}{\em Individual Phrases} &
\multicolumn{2}{c|}{\em Pair of Phrases} \\
\hline
\multicolumn{1}{|c|}{\em Attribute} &
\multicolumn{1}{c|}{\em Value} &
\multicolumn{1}{c|}{\em Attribute} &
\multicolumn{1}{c|}{\em Value} \\
\hline
NAME-1        & YES &
ALIAS         & NO
\\
JV-CHILD-1    & NO &
BOTH-JV-CHILD & NO
\\
NAME-2        & YES &
COMMON-NP     & NO
\\
JV-CHILD-2    & NO &
SAME-SENTENCE & NO
\\
\hline
\end{tabular}
\end{small}
\caption{Attributes and Values for EJV \object{entity} instance.}
\label{table:ejv-instance}
\end{table}

Of the 8 features used in this experiment, two focus on the first
reference, two focus on the second reference and four are based on
the pair of references.  The following is a brief description of the
features that focus on individual phrases, where $i \in \{1,2\}$.

\begin{itemize}

\item{\feature{NAME-$i$:}}
Does reference $i$ contain a name?
Possible values: \{{\sc yes, no}\}.

\item{\feature{JV-CHILD-$i$:}}
Does reference $i$ refer to a joint venture child, i.e., a company
formed as the result of a tie-up among two or more \object{entities}?
Possible values: \{{\sc yes, no, unknown}\}.

\end{itemize}

The last four features focus on the pair of references.

\begin{itemize}

\item{\feature{ALIAS:}}
Does one reference contain an alias of the other, i.e., does each
reference contain a name and is one name a substring of the other
name?\footnote{Note that some texts contain more than one entity for
which a given name might be an alias under this definition, e.g.,
``SUMITOMO'' is a substring of both ``SUMITOMO CORP.'' and ``SUMITOMO
ELECTRICAL INDUSTRIES LTD.'', so this feature is not always a reliable
indicator of coreference.}
Possible values: \{{\sc yes, no}\}.

\item{\feature{BOTH-JV-CHILD:}}
Do both references refer to a joint venture child?  This feature is
defined as

\begin{description}
\item[yes] when $\forall i, \mbox{\sc jv-child-{\em i} = yes}$
\item[no] when  $\forall i, \mbox{\sc jv-child-{\em i} = no}$
\item[unknown] otherwise.
\end{description}

\item{\feature{COMMON-NP:}}
Do the references share a common noun phrase?  Some references contain
non-simple noun phrases, e.g., appositions and relative clauses.  This
feature compares the simple constituent noun phrases of each
reference.  Possible values: \{{\sc yes, no}\}.

\item{\feature{SAME-SENTENCE:}}

Do the references come from the same sentence?  \systemname{Resolve}\ does not
use
\systemname{circus}\ output, and thus has no notion of a trigger
family as it was used in the \muc{5} system; the {\sc same-sentence}
feature is a very weak attempt to extract this sort of information.
Possible values: \{{\sc yes, no}\}.

\end{itemize}

\begin{figure}
\centering
\hspace*{\fill} \epsfxsize=2.8in
\epsffile{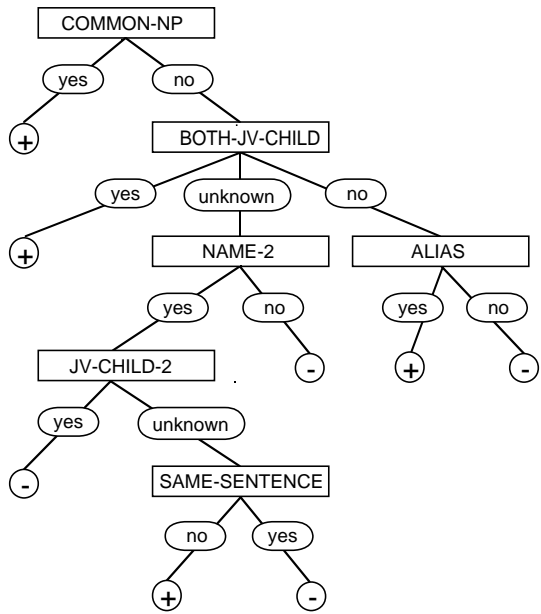}
\hspace*{\fill}

\caption{A pruned C4.5 decision tree}
\label{fig:pruned-tree}
\end{figure}

1230 feature vectors, or instances, were created from the
\object{entity} references marked in the 50 texts.  Of these, 322
(26\%) were {\em positive} (``+'') instances -- pairs of phrases that
were coreferent -- and 908 (74\%) were {\em negative} (``-'')
instances -- pairs of phrases that were not coreferent.  Figure
\ref{fig:pruned-tree} shows a pruned C4.5 decision tree trained on all
the instances.

\subsection{Evaluation Methodology}
\label{evaluation}

Coreference is a symmetrical and transitive relation that holds among
a set of two or more references, e.g., if we know that {\em A} is
coreferent with {\em B}, and {\em B} is coreferent with {\em C}, then
there is an implicit coreference ``link'' between {\em A} and {\em
C}.\footnote{As was noted earlier, some references are {\em
multireferent}, i.e., they have more than one referent.  Thus, if {\em
B} is multireferent, we cannot conclude that {\em A} is coreferent
with {\em C}; for example, if {\em A = Sneezy}, {\em B = the dwarfs}
and {\em C = Grumpy}, we don't want to infer that {\em Sneezy} = {\em
Grumpy}.  We can ignore such complications in this paper since the
experiments reported herein exclude multireferent phrases.}  Any
coreference classification for two references has implications beyond
the determination of whether that particular classification was
correct or incorrect.  For example, if {\em A} and {\em B} are
correctly classified as coreferent, but {\em B} and {\em C} are
incorrectly classified as not coreferent, a system may also
incorrectly conclude that {\em A} and {\em C} are not coreferent.
Thus, simply measuring the accuracy of a coreference classifier is
inadequate for evaluating how well the classifier performs its task.

Two metrics that have been used to evaluate the performance of IE
systems are {\em recall} and {\em precision}
\cite{muc3:intro:metrics,muc4:intro:metrics,muc5:intro:metrics}.
Recall is the percentage of information in a text that is correctly
extracted by a system; precision is the percentage of information
extracted by a system that is correct.  For example, if a text
contains four relevant items (represented by \{{\em A, B, C, D\/}\} in
an answer {\em key}), and a system correctly extracts the three items
\{{\em A, B, C\/}\} but incorrectly extracts the two additional items
\{{\em E, F\/}\} (represented by \{{\em A, B, C, E, F\/}\} in a system
{\em response}), then its recall would be 75\% and its precision would
be 60\%.

A function to combine recall and precision into a single measure of
performance was incorporated into the Fourth Message Understanding
Evaluation and Conference \cite{muc4:intro:metrics}.  The {\em
F-measure}, a metric used to evaluate Information Retrieval (IR)
system performance \cite{vanrijsbergen:79:ir}, combines recall and
precision scores into a single number using the formula
\[
F =
\frac{(\beta^{2} + 1.0) \times P \times R}
     {\beta^{2} \times P + R}
\]
where {\em P} is the precision score, {\em R} is the recall score and
$\beta$ is the relative weight given to recall over precision.  For
example, a $\beta$ value of 1.0 gives equal weight to recall and
precision; a value of 2.0 gives recall twice the weight of precision;
a value of 0.5 gives recall half the weight of precision.

An evaluation methodology for the coreference task is being developed
for the upcoming Sixth Message Understanding Evaluation and Conference
(\muc{6}).  The metrics used for evaluating overall IE system
performance are being adapted for use on this subtask
(cf. \cite{mitre:94:coref-scoring}), where the answer key for each
text contains a set of phrases and the coreference links among them.
However, evaluation of coreference performance is complicated by the
need to take into account the {\em implicit} coreference links among
phrases.  Thus, transitive closures are taken for both the answer
key (the {\em key closure}) and the system response (the {\em response
closure}).  Recall is measured by the percentage of explicit
coreference links in the key that are also found in the response
closure, i.e., what fraction of correct coreference links is implied
by the transitive closure of the coreference links in the system
response.  Precision is measured by the percentage of explicit
coreference links in the response that are also found in the key
closure, i.e., what fraction of coreference links in the response is
implied by the transitive closure of the coreference links in the key.

\subsection{Results}
\label{results}

One experiment was run using \systemname{resolve}.  In this experiment, for
each
set of instances taken from the 50 texts, one set was selected for
testing purposes and the remaining sets were used to train a new
decision tree.  This process was iterated over all 50 sets of
instances.  The results shown in Table \ref{table:results} represent
the average of these iterations: the first row shows the recall,
precision and F-measure ($\beta = 1.0$) scores for unpruned decision
trees; the second row shows the results for pruned decision
trees.\footnote{Default settings for all C4.5 parameters were used
throughout this experiment (see \cite{quinlan:93:c4.5}, Chapter 9, for
more information about C4.5 parameters).}

The third row in Table \ref{table:results} shows the results from a
second experiment, in which the rule set from the coreference module
of the UMass/\-Hughes \muc{5} IE system was applied to the memory token
pairs generated from the references marked using \systemname{cmi}.

\begin{table}
\centering
\begin{small}
\begin{tabular}{|l|c|c|c|} \hline
System               & Recall & Precision & F-measure \\ \hline \hline
\systemname{Resolve}\ (unpruned) & 85.4\% & 87.6\%    & 86.5\% \\ \hline
\systemname{Resolve}\ (pruned)   & 80.1\% & 92.4\%    & 85.8\% \\ \hline
\muc{5} rule set     & 67.7\% & 94.4\%    & 78.9\% \\ \hline
\end{tabular}
\end{small}
\caption{Results for EJV \object{entity} coreference resolution.}
\label{table:results}
\end{table}

\subsection{Discussion}
\label{discussion}

When we first began applying decision trees to the coreference
resolution problem, we were hoping to achieve performance that was
comparable to the manually engineered rules we had used in \muc{5}.
We were greatly encouraged to discover that we could achieve
performance that surpassed the performance of the rules from our
\muc{5} system in both recall and F-measure scores.

As was noted earlier, the \muc{5} coreference rules were designed to
minimize false positives.  The effect of this bias can be seen in the
higher precision score achieved by the rule set in comparison with
both the unpruned and pruned decision trees.  The difference in
precision scores between the unpruned and pruned versions of the
decision trees might be explained by the prevalence of negative
instances (74\%) in the data set, which may lead to a stronger bias to
classify pairs of phrases as not coreferent in the smaller trees.

The comparative effects of false positives and false negatives in
coreference classification on overall IE system performance remains an
open question.  However, while the precision scores achieved by the
decision trees and the rule-base are rather close, especially for the
pruned version of the trees, there is a large difference between their
recall scores.  Until we can ascertain the relative importance of high
recall vs. high precision in overall IE system performance, the
F-measure score that gives equal weight to recall and precision may be
the best indicator of overall performance on the coreference
resolution task.  However, as can be seen in Table
\ref{table:results-f-measures}, when \systemname{resolve}\ uses pruning, its
performance surpasses that of the rule set even when the recall score
is given twice the weight of precision score or when the recall score
is given half the weight of precision score.\footnote{The pruned
decision trees yield higher F-measure scores than the \muc{5} rule
set unless the recall score is given less than one-third the weight of
the precision score.}

\begin{table}
\centering
\begin{small}
\begin{tabular}{|l|c|c|c|} \hline
System               & $\beta= 2.0$ & $\beta = 1.0$ & $\beta = 0.5$ \\ \hline
\hline
\systemname{Resolve}\ (unpruned) & 85.8\% & 86.5\% & 87.1\% \\ \hline
\systemname{Resolve}\ (pruned)   & 82.3\% & 85.8\% & 89.6\% \\ \hline
\muc{5} rule set     & 71.8\% & 78.9\% & 87.5\% \\ \hline
\end{tabular}
\end{small}
\caption{F-measures for different values of $\beta$.}
\label{table:results-f-measures}
\end{table}

\section{Conclusions}
\label{conclusion}

One of the original goals of this new approach was to develop a system
that achieved good performance in resolving references -- performance
that was at least as good as the performance achieved using manually
engineered rules in our \muc{5} system.  However, as we continue to
pursue this approach, we find that there is another advantage to using
decision trees: they allow us to focus on determining which features
work well for resolving references.

We are encouraged by the performance of the decision trees on the
coreference resolution problem.  The features we have used in the
experiment described above are not considered comprehensive by any
means.  While they have proved sufficient for attaining a certain
level of performance, an examination of specific errors made by the
trees shows that additional features will be needed to attain higher
levels.

One area we will develop further is a set of features that incorporate
syntactic knowledge.  We don't have any features that identify the
various syntactic constituents of a sentence, e.g., subject or direct
object, nor do we have any features that identify clause boundaries
(only sentence boundaries).  These features will be incorporated in
future experiments.  Features based on focus of attention
\cite{sidner:79:anaphora,grosz:83:def-np}, which presuppose knowledge
about syntactic constituents may also prove useful.  Our experiment
used a feature set that was largely semantic in nature: it is
interesting to see how well semantic features work as a basis for
coreference resolution ... and it is not surprising to see that they
are also insufficient.

Ultimately, we hope to understand better which features are important
for coreference classification, across different objects and different
domains.  Such an understanding would benefit people involved with IE
system development, and should be of interest to people outside the IE
community as well.  We think that decision trees are an important tool
in a systematic study of coreference resolution.

\bibliographystyle{named}

\setlength\itemsep{0pt}

\end{document}